\documentclass[
]{ceurart}

\usepackage{listings}
\usepackage{mathptmx}
\usepackage{graphicx}
\usepackage{xcolor}
\newcommand{\sq}[1]{`#1'}

\newcommand{\dqi}[1]{``\emph{#1}''}
\newcommand{\qdfull}[2]{\emph{\textbf{#1}}~(\texttt{#2})}
\newcommand{\qdn}[1]{\emph{\textbf{#1}}}
\newcommand{\qd}[1]{\texttt{#1}}  
\newcommand{\con}[1]{\emph{\textbf{#1}}}

\begin{document}

\copyrightyear{2025}
\copyrightclause{Copyright for this paper by its authors.
  Use permitted under Creative Commons License Attribution 4.0
  International (CC BY 4.0).}

\conference{AIC 2025: The 10th International Workshop on Artificial Intelligence and Cognition (held as part of ECAI 2025). October 25-26, 2025. Bologna, Italy}

\title{Abstract Concept Modelling in Conceptual Spaces: \\A Study on Chess Strategies}

\author[1]{Hadi Banaee}[%
orcid=0000-0002-9607-9504,
email=hadi.banaee@oru.se,
]
\cormark[1]
\address[1]{Department of Computer Science, \"Orebro~University, Sweden}

\author[1]{Stephanie Lowry}[%
orcid=0000-0003-3788-499X,
email=Stephanie.Lowry@oru.se,
]

\cortext[1]{Corresponding author.}

\begin{abstract}
We present a conceptual space framework for modelling abstract concepts that unfold over time, demonstrated through a chess-based proof-of-concept. 
Strategy concepts, such as attack or sacrifice, are represented as geometric regions across interpretable quality dimensions, with chess games instantiated and analysed as trajectories whose directional movement toward regions enables recognition of intended strategies. This approach also supports dual-perspective modelling, capturing how players interpret identical situations differently.
Our implementation demonstrates the feasibility of trajectory-based concept recognition, with movement patterns aligning with expert commentary.
This work explores extending the conceptual spaces theory to temporally realised, goal-directed concepts. 
The approach establishes a foundation for broader applications involving sequential decision-making and supports integration with knowledge evolution mechanisms for learning and refining abstract concepts over time.
 
\end{abstract}


\maketitle

\section{Introduction}

In the 1956 \sq{Game of the Century,} Fischer sacrificed his queen to launch a decisive attack against Byrne, moves that defied immediate evaluation but reflected deep intentional strategy. 
How can such intended strategies be computationally recognised from the unfolding sequence of board positions?


Human strategy-making is a form of abstract cognition where meaning emerges not from isolated actions but from sequences that unfold toward latent objectives. These abstract concepts, such as attack preparation or positional sacrifice, pose significant challenges for traditional knowledge representation systems. 
They are not statically defined, and their realisation depends on temporal coherence and contextual interpretation. Moreover, most existing AI systems in domains like chess or human behaviour modelling rely on predictive engines or symbolic rules, lacking the structural interpretability and temporal grounding needed to capture such conceptual patterns.

Conceptual spaces theory offers a geometric framework for representing abstract knowledge as regions in interpretable multi-dimensional spaces. 
It enables similarity-based reasoning and captures structural relationships between concepts.
While successfully applied to perceptual and semantic domains, its potential for modelling temporally realised, intentional constructs remains largely unexplored. We argue that strategy concepts can be modelled as geometric regions and recognised through temporal movement patterns in conceptual space.

We propose a conceptual space framework where chess strategies are defined as regions across interpretable quality dimensions such as pressure, control, and material balance. 
By embedding game positions as points in this space and analysing how sequences of moves form trajectories, we explore a novel method for identifying high-level strategies through temporal geometric strategy recognition.
Our framework supports dual-perspective modelling, generating distinct trajectories to capture player-relative interpretations.
Chess provides an ideal testbed with well-documented strategy concepts, rich temporal sequences, and expert annotations, facilitating the validation opportunities for this approach.

This paper presents a proof-of-concept demonstrating the feasibility of: 
(1) representing strategy concepts as multi-domain geometric regions in chess, 
(2) recognising intentional strategies by integrating interpretable dimensions, region structures, and trajectory-based analysis, 
and 
(3) proposing an extension of the conceptual spaces theory to 
model goal-directed abstract concepts in interpretable AI systems.

\section{Background and Related Work}

\subsection{Conceptual Spaces Theory}

\emph{Conceptual spaces theory}, introduced by Gärdenfors~\cite{gardenfors2000conceptual}, provides a geometric framework for representing knowledge. In this model, concepts are defined as convex regions in a multi-dimensional space, where each dimension corresponds to a cognitively meaningful and interpretable quality (e.g., temperature, hue, weight). These dimensions are often grouped into domains, such as colour or taste, enabling structured representations of complex concepts.

A key strength of conceptual spaces is their support for \emph{natural similarity metrics}. This enables graded category membership and prototype-based reasoning: items near the region centre are considered more typical, while those near the boundaries are less so~\cite{gardenfors2000conceptual, gardenfors2014geometry}. This prototype effect aligns well with human categorisation, making conceptual spaces cognitively plausible. 
The framework has been successfully applied in perceptual and sensory domains, including spatial reasoning and object grouping~\cite{zenker2015applications, lieto2017conceptual, osta2023reasoning}. These examples demonstrate its cognitive plausibility and interpretability. 

However, despite its strengths, constructing conceptual spaces presents challenges, particularly in identifying appropriate quality dimensions and anchoring meaningful concept regions~\cite{banaee2018datadriven}. This is especially difficult in domains involving \emph{abstract concepts}, where relevant dimensions are not directly perceptual or physical. For example, strategies such as \con{Space~Domination} in chess lack clear sensory referents, making dimension construction non-trivial.

To date, conceptual spaces have been successfully applied in perceptual domains; however, their use in modelling abstract, temporal, or goal-oriented concepts remains limited. 
%
While conceptual spaces support \emph{concept learning} through region adjustment~\cite{douven2024concept}, their potential for modelling dynamic, evolving concepts has yet to be fully realised. This gap motivates our extension of the theory to model abstract concepts, using chess as a foundational testbed to demonstrate feasibility and develop methodological approaches.

\subsection{Chess Strategy in AI Systems}

\emph{Chess strategy} concerns long-term planning and the pursuit of abstract goals such as structural advantages and control of key areas~\cite{nimzowitsch1925my, silman2010how}. However, the chess literature lacks systematic formalisation of concepts, often conflating strategy with tactical motifs or describing strategies using intuitive language~\cite{watson1998secrets}. 

Computational models of chess have traditionally focused on position evaluation rather than modelling abstract intentions. 
Early chess systems relied on handcrafted rules and pattern databases, including material evaluation, piece-square tables, and symbolic features like king safety or centre control. Search-based engines like Stockfish used minimax and alpha-beta pruning, guided by knowledge-engineered heuristics that encode static preferences \cite{maharaj2021chessai}.

Modern engines such as AlphaZero~\cite{silver2017mastering} and LeelaChessZero~\cite{jenner2024advances} learn strategic patterns via deep reinforcement learning from self-play. These systems develop policy-value networks to predict move probabilities and position values without relying on human chess knowledge. While powerful, such models encode strategy implicitly and lack interpretability or transparent reasoning~\cite{mcgrath2022acquisition}.

Recent work has begun addressing interpretability challenges in chess AI. DecodeChess~\cite{decodechess2022} 
integrates engine evaluations into symbolic human-aligned motifs such as threats, plans, and tactical motifs~\cite{reid2020aligning}. 
%
%
Additionally, explainable AI (XAI) techniques have been used to uncover internalised chess strategies within neural chess agents~\cite{bjornsson2024chess}. Hammersborg and Strümke propose a method for chess to uncover concept-activated reasoning within open-source neural chess models, revealing which information the model uses~\cite{hammersborg2024information}. The Maia is another engine that predicts human decisions at various skill levels, offering insights into human-like tendencies~\cite{tang2024maia}. LLM-based systems attempt to extract notion-level vectors from expert-like models and generate explanations informed by interpretable reasoning~\cite{kim2025bridging}.

Despite these advances, current approaches still emphasise move-level evaluation and prediction rather than modelling abstract concepts as structured, interpretable entities that evolve across time. While explainable AI methods have been applied to chess agents, no existing approach represents strategies as regions grounded in an interpretable multidimensional space or tracks strategic intention through trajectory-based reasoning. 
This paper addresses this gap by proposing a conceptual spaces framework that models chess strategies as geometric regions and recognises intentions through temporal movement patterns toward abstract goals.

\section{Chess Conceptual Space: Design and Implementation}
\label{sec:chess-cs}
To explore how abstract and temporally evolving concepts can be modelled in interpretable geometric spaces, we focus on chess as a domain rich in explicit strategy constructs, expert knowledge, and structured game data.
This section presents our proposed design and modules of a chess-specific conceptual space, illustrating how interpretable dimensions, concepts, and trajectories can support the recognition and interpretation of strategies. 
While this implementation focuses on chess, the methodology reflects a more general approach to modelling abstract concepts, which we revisit in Section~\ref{sec:general&discuss}.

\subsection{Knowledge Acquisition: Abstract Strategy Concepts in Chess}

To construct a conceptual space grounded in chess strategy understanding, we gathered abstract strategy concepts and interpretable quality dimensions. Our methodology focused on identifying concepts that are temporally extended, multi-faceted, and traceable through gameplay sequences. We began with McGrath et al.'s analysis of concept acquisition in AlphaZero~\cite{mcgrath2022acquisition}, which formalises primitive chess elements such as material imbalances, positional pressure, and tactical motifs, providing a computational foundation for strategy understanding.
%
To extend this foundation, we analysed annotated games from chess databases, consulted classical strategy texts, and examined online educational content. 
We also used large language models to explore chess strategies from available sources such as expert blogs and educational platforms, generating an initial list of candidate strategies. The retrieved candidate concepts were manually verified and filtered based on two criteria: temporal observability (concepts that develop over multiple moves) and computational measurability (concepts that can be quantified from position features).

From this process, we compiled an initial pool of strategy concepts and quality dimensions, which we subsequently filtered to three core strategies as abstract concepts for implementation: \con{King~Attack}, \con{Positional~Sacrifice}, and \con{Space~Domination}. These strategies are chosen based on their strategy richness, observability over time, and contrasting patterns across multiple conceptual dimensions. These selections provide suitable exemplars for testing the representational capacity of our geometric framework.

\subsection{Space Construction: Dimension Grounding and Concept Anchoring}
To build a geometric space for abstract strategy concepts in chess, we grounded a set of interpretable quality dimensions that capture primitive yet computationally measurable factors involved in chess strategic reasoning. 
Our goal was to select dimensions that are both strategically meaningful and measurable from game positions. In the proposed framework, we employed seven quality dimensions: 
\qdfull{Material}{MAT}, 
\qdfull{Mobility}{MOB}, 
\qdfull{Vulnerability}{VUL}, 
\qdfull{Control}{CTR}, 
\qdfull{Flow}{FLO}, 
\qdfull{Pressure}{PRS}, and 
\qdfull{Space}{SPA}. 
Each dimension is normalised for geometric consistency.
We organised these dimensions into three semantic domains based on their relationships in chess understanding, where each domain combines dimensions that co-occur in chess patterns and reflect decision-making aspects.
\begin{itemize}
    \item \textbf{Territory Domain} (\qd{CTR}, \qd{FLO}) captures \dqi{How efficiently do I control and coordinate?}, linking central control with coordination of piece movement;
    \item \textbf{Force Domain} (\qd{MAT}, \qd{MOB}, \qd{SPA}) represents \dqi{What power and activity do I have?}, combining material resources and piece activity with spatial advantages;
    \item \textbf{Conflict Domain} (\qd{PRS}, \qd{VUL}) addresses \dqi{What threats and risks exist?}, balancing immediate pressure with risk exposure.
\end{itemize}


We anchored our three strategy concepts as convex regions within this multi-dimensional space using expert chess knowledge. Importantly, these regions span multiple domains, ensuring that no concept is reducible to a single metric. 
%
For example, \con{King~Attack} involves high \qdn{pressure} (actively threatening the enemy king) and low \qdn{vulnerability} (maintaining king safety while attacking) with variable \qdn{material}, as it often relies on active threats and accepts defensive risk or sacrifices. In contrast, \con{Space~Domination} emerges from enhanced \qdn{control} and \qdn{space} advantage (occupying key squares and territory) without immediate tactical engagement, reflecting a long-term, positional approach.

\begin{figure}
    \centering
    \includegraphics[width=0.99\linewidth]{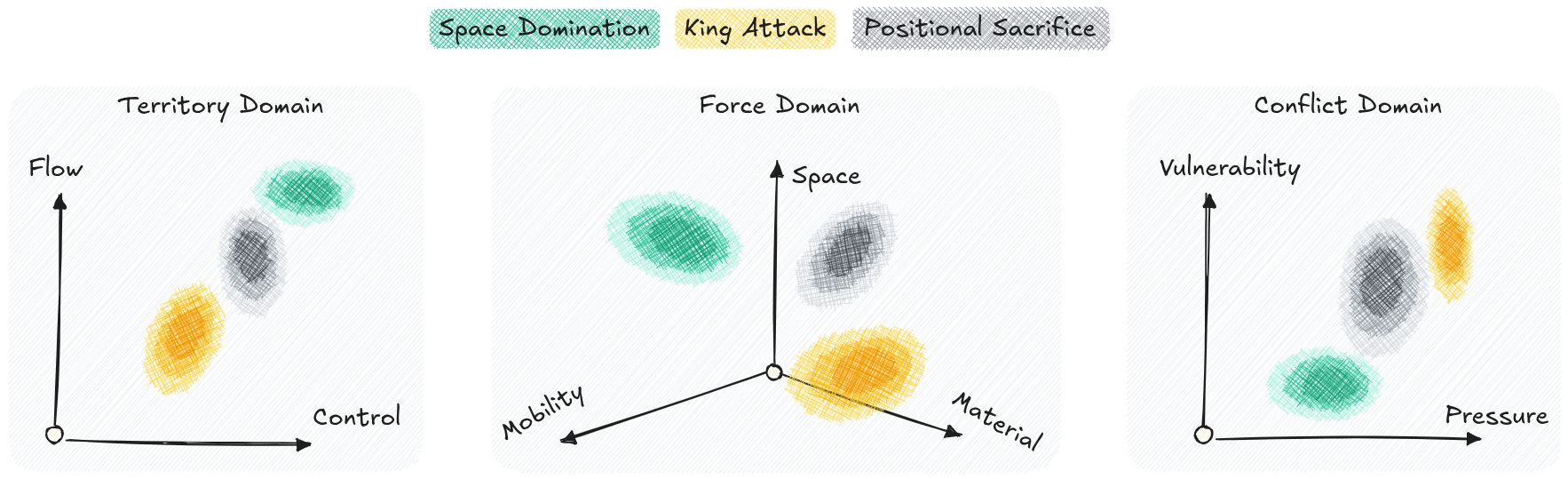}
    \caption{An illustration of the chess conceptual space, showing chess strategies anchored as regions across force, territory, and conflict domains.}
    \label{fig:chess_space}
\end{figure}

These qualitative characterisations are anchored to specific quantified boundaries within our space, where each concept is defined as a set of convex regions distributed across domains. 
Figure~\ref{fig:chess_space} illustrates our proposed chess conceptual space, showing how abstract strategies are embedded as concept regions across the domains. 
Note that in the current implementation of the framework, the region boundaries were defined heuristically based on chess principles and expert strategic knowledge. 

\subsection{Space Instantiation: Position Encoding and Game Trajectories}

The conceptual space becomes operational by populating it with real chess positions. To instantiate our conceptual space, we embed chess positions as multi-dimensional points in the proposed space and map complete games as temporal trajectories capturing strategy evolution.


A chess position is transformed into a normalised multi-dimensional point through computational analysis across our quality dimensions. Using \emph{python-chess}\footnote{https://python-chess.readthedocs.io/en/latest/} 
library functions, we compute material balance (difference in piece values), mobility (number of legal moves available), control (influence over central squares \texttt{e4}, \texttt{d4}, \texttt{e5}, \texttt{d5}), pressure (weighted attacks against opponent pieces), and spatial advantage (pieces advanced into opponent territory). For instance, \qdfull{Pressure}{PRS} combines value-weighted attacks against all enemy pieces, while \qdfull{Control}{CTR} measures influence over the four central squares (\texttt{e4}, \texttt{d4}, \texttt{e5}, \texttt{d5}). This systematic transformation ensures geometric consistency while preserving interpretability.

To generate game-specific trajectories, game sequences $G = {p_1, p_2, ..., p_n}$ are converted into conceptual space trajectories $T = {\mathbf{v}_1, \mathbf{v}_2, ..., \mathbf{v}_n}$ where each move transition corresponds to geometric movement $\mathbf{v}i \rightarrow \mathbf{v}{i+1}$. This enables temporal analysis of concept development beyond static position evaluation. 

\begin{figure}
    \centering
    \includegraphics[width=0.95\linewidth]{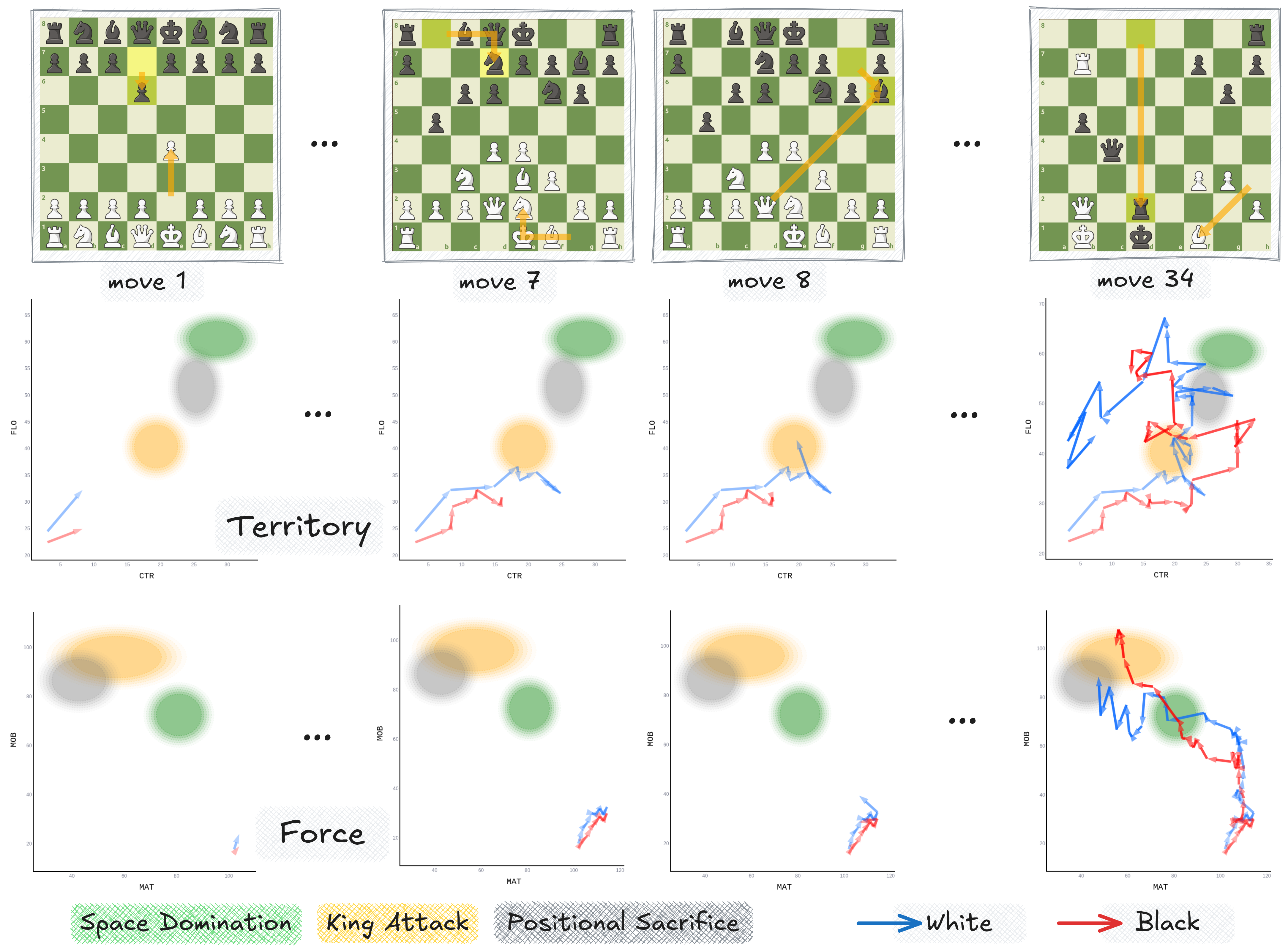}
    \caption{Dual-perspective trajectories of chess board positions in chess conceptual space.
}

    \label{fig:trajectories}
\end{figure}

Our framework generates separate trajectories from \textit{White} and \textit{Black} strategy viewpoints for each game. While board positions remain identical, quality dimension values reflect player-specific interpretations. This dual representation captures subjective aspects of strategy reasoning. Figure~\ref{fig:trajectories} demonstrates this through a move sequence example from a master game\footnote{https://www.chess.com/analysis/game/pgn/4uQNJeFLLv/analysis}, 
showing how identical positions generate distinct interpretations across \textit{Territory} and \textit{Force} domains.

\subsection{Strategy Recognition and Interpretation Examples}
Based on our proposed chess conceptual space, we explore strategy recognition through geometric region membership analysis. Strategic alignment is defined via trajectory movement analysis within predefined concept regions $R_k$, where movement converges toward region centroids. For a trajectory $T = \{\mathbf{v}_1, \mathbf{v}_2, ..., \mathbf{v}_n\}$, a concept $C_k$ is identified when the recent trajectory segment $T_{i:j}$ satisfies: $\mathbf{v}_t \in R_k$ for consecutive moves and $\text{direction}(T_{i:j}) \rightarrow \text{centre}(R_k)$. This formulation distinguishes sustained strategic progression from incidental fluctuations, ensuring both regional continuity and directional alignment.

\begin{figure}
    \centering
    \includegraphics[width=0.85\linewidth]{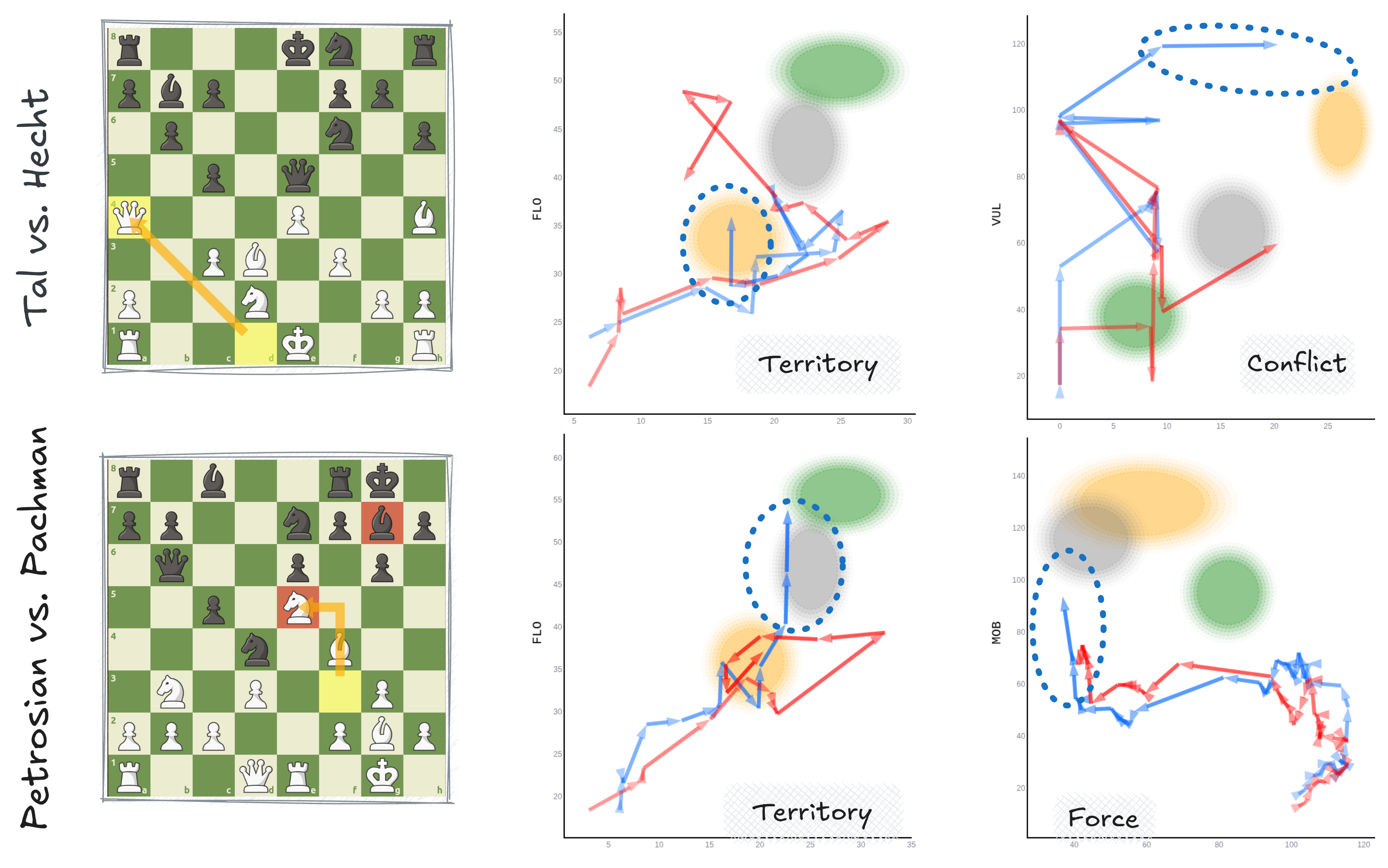}
    \caption{Strategy recognition examples aligned with concept regions in annotated master games.}
    \label{fig:recog}
\end{figure}

We demonstrate recognition capabilities through two annotated master games. In Tal vs. Hecht (1962), White's trajectory approaches the \con{King~Attack} region spanning the Territory and Conflict domains after move 15, remaining within region boundaries for three consecutive moves while moving toward the concept centroid. Expert annotations confirm tactical assault during this movement. In Petrosian vs. Pachman (1961), White's trajectory approaches the \con{Positional~Sacrifice} region across the Territory and Force domains following an exchange sacrifice, maintaining region membership over two moves after move 13. The geometric pattern aligns with expert commentary describing positional compensation for material investment. Figure~\ref{fig:recog} visualises these trajectory patterns and their correspondence with the anchored concept regions, shown only on the Force domain.


Our framework generates separate classification results for White and Black perspectives from identical board sequences. Moreover, recognised strategies are triggered across domain combinations rather than isolated dimension changes. For example, \con{King~Attack} recognition requires simultaneous presence in both the Force domain (with high \qd{MOB}) and the Conflict domain (with high \qd{PRS} and \qd{VUL}), while \con{Positional~Sacrifice} requires entry into the Territory domain (\qd{CTR}, \qd{FLO}) along with declining \qd{MAT} in the Force domain. This asymmetric recognition demonstrates that strategy interpretations are player-relative, capturing competing intentions from identical positions. Additionally, the multi-domain requirement reduces false positives from single-dimension fluctuations.

This recognition mechanism is exploratory and heuristic, but it shows alignment between geometric movement and human-understood strategic intent. Recognition depends on consistent multi-move region presence and directional convergence. 

\section{Discussion}
\label{sec:general&discuss}

\subsection{Generalisation and Framework Extensions}

\begin{figure}
    \centering
    \includegraphics[width=0.99\linewidth]{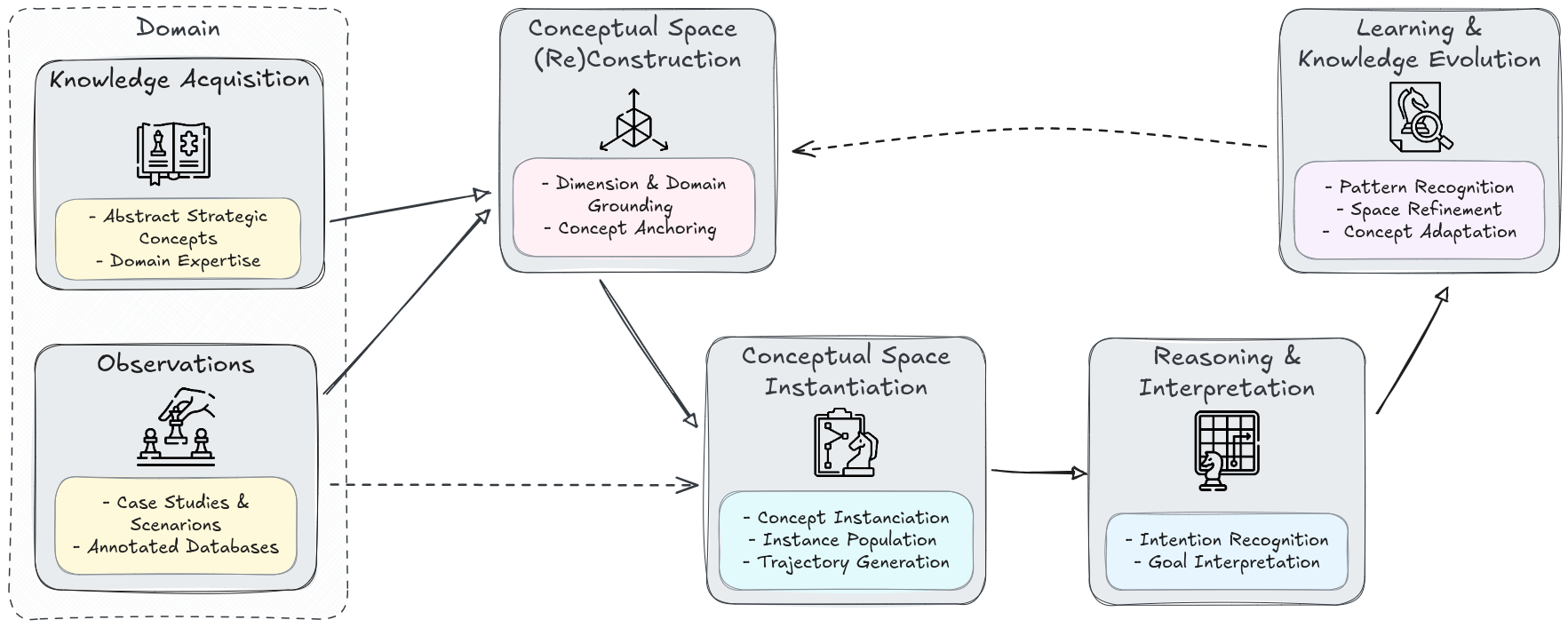}
    \caption{Proposed framework architecture for geometric modelling of abstract concepts.}
    \label{fig:fr_diagram}
\end{figure}

The chess implementation demonstrates components of a broader methodology for geometric modelling of abstract concepts. 
Figure~\ref{fig:fr_diagram} illustrates the broader framework from which our chess system is derived.  The design enables abstract concepts to be anchored in interpretable quality dimensions, embedded as geometric regions, and tracked over time via trajectory analysis. 
%
While our proposed chess conceptual space provides a proof-of-concept implementation for trajectory-based reasoning, modules, such as adaptive learning (automatically refining concept boundaries from game data) and knowledge evolution (discovering new strategy concepts over time), remain central to the full potential of the framework.

Two critical modules, pattern recognition and adaptive learning, remain unimplemented yet represent essential theoretical advances for conceptual spaces research. The knowledge evolution capability addresses a fundamental limitation.
Our framework enables spaces to evolve through observation and concept boundary refinement, allowing them to adapt to emerging patterns and novel concepts.

The framework generalises to domains involving goal-directed, sequential decision-making where abstract concepts manifest through temporal action sequences. Examples include social interaction dynamics, collaborative planning, narrative structure understanding, and multi-agent coordination. The approach aligns with spatially grounded human cognition while enabling computational operationalisation of abstract concepts that lack direct perceptual grounding but emerge through behavioural observation and strategic interpretation.

\subsection{Challenges and Implications}

Quality dimensions and concept regions are defined heuristically through expert knowledge, without data-driven discovery mechanisms. This expert-dependent construction limits scalability and introduces potential subjective bias in dimensional selection and region boundary definition.

Current validation remains qualitative and limited to individual annotated master games. While trajectory patterns align with expert strategic commentary, systematic evaluation across large game corpora is required to establish reliability and generalisability claims. 

These limitations reflect broader methodological challenges in geometric modelling of abstract concepts, particularly their temporal evolution and subjective interpretation. The problem of automatically discovering quality dimensions for abstract concepts, learning concept boundaries from observational data, and validating geometric representations against human expert knowledge represents significant open research questions for both artificial intelligence and conceptual spaces theory.

\section{Conclusion and Future Directions}

This paper introduced a conceptual space framework for modelling abstract concepts, demonstrated through a chess-based proof-of-concept. By grounding interpretable quality dimensions, anchoring chess strategies as geometric regions, and analysing game trajectories as movement patterns, we demonstrated a novel approach to chess strategy recognition and interpretation.
The chess implementation confirms the feasibility of trajectory-based recognition and dual-perspective modelling, with promising alignment to expert annotations. Key innovations include geometric intention recognition through directional movement toward concept regions and an attempt towards extension of the conceptual spaces theory to temporal, goal-oriented abstract concepts.

Several directions for future research emerge from this foundation. Immediate priorities include systematic validation of chess conceptual space through large-scale annotated game databases. Data-driven approaches for learning concept regions and quality dimensions from annotated databases would reduce expert dependency and improve scalability. 
These developments will refine the framework into a practical tool for interpretable strategy analysis and open pathways for its adaptation to domains such as multi-agent planning and human-AI coordination, where evolving conceptual patterns play a central role.

\begin{acknowledgments}
  This work was supported by the Swedish Knowledge Foundation in the TeamRob Synergy Project (contract number 20210016). 
\end{acknowledgments}

\bibliography{references}

@String{Computing = "Computing" }

@String{Springer = "Springer-Verlag" }

@book{gardenfors2000conceptual,
title = "Conceptual spaces : the geometry of thought",
author = "Peter G{\"a}rdenfors",
year = "2000",
language = "English",
isbn = "9780262572194",
series = "A Bradford book",
publisher = "MIT Press",
}

@book{gardenfors2014geometry,
title = "Geometry of meaning : semantics based on conceptual spaces",
author = "Peter G{\"a}rdenfors",
year = "2014",
language = "English",
isbn = "9780262026789",
publisher = "MIT Press",
}

@book{zenker2015applications,
  editor = {Zenker, Frank and Gärdenfors, Peter},
  title = {Applications of Conceptual Spaces: The Case for Geometric Knowledge Representation},
  year = {2015},
  publisher = {Springer Cham},
  series = {Synthese Library},
  volume = {359},
  doi = {10.1007/978-3-319-15021-5},
  isbn = {978-3-319-15021-5} 
}

@article{lieto2017conceptual,
  author  = {Antonio Lieto and Antonio Chella and Marcello Frixione},
  title   = {Conceptual Spaces for Cognitive Architectures: A Lingua Franca for Different Levels of Representation},
  journal = {Biologically Inspired Cognitive Architectures},
  volume  = {19},
  pages   = {1--9},
  year    = {2017}
}

@article{douven2024concept,
title = "Concept Learning: Convexity Versus Connectedness",
author = "Igor Douven and Steven Verheyen",
year = "2024",
month = dec,
day = "9",
doi = "10.1007/s10670-024-00909-1",
language = "English",
journal = "Erkenntnis",
issn = "0165-0106",
publisher = "Springer Netherlands",
}

@article{banaee2018datadriven,
author = {Banaee, Hadi and Schaffernicht, Erik and Loutfi, Amy},
title = {Data-driven conceptual spaces: creating semantic representations for linguistic descriptions of numerical data},
year = {2018},
issue_date = {September 2018},
publisher = {AI Access Foundation},
address = {El Segundo, CA, USA},
volume = {63},
number = {1},
issn = {1076-9757},
doi = {10.1613/jair.1.11258},
journal = {J. Artif. Int. Res.},
month = sep,
pages = {691–742},
numpages = {52}
}

@article{osta2023reasoning,
  author    = {Peter G{\"a}rdenfors and Mat{\'i}as Osta-V{\'e}lez},
  title     = {Reasoning with Concepts: A Unifying Framework},
  journal   = {Minds and Machines},
  year      = {2023},
  doi       = {10.1007/s11023-023-09640-2}
}

@article{maharaj2021chessai,
  author       = {Shiva Maharaj and
                  Nick Polson and
                  Alex Turk},
  title        = {Chess {AI:} Competing Paradigms for Machine Intelligence},
  journal      = {CoRR},
  volume       = {abs/2109.11602},
  year         = {2021},
}

@article{silver2017mastering,
  author       = {David Silver and
                  Thomas Hubert and
                  Julian Schrittwieser and
                  Ioannis Antonoglou and
                  Matthew Lai and
                  Arthur Guez and
                  Marc Lanctot and
                  Laurent Sifre and
                  Dharshan Kumaran and
                  Thore Graepel and
                  Timothy P. Lillicrap and
                  Karen Simonyan and
                  Demis Hassabis},
  title        = {Mastering Chess and Shogi by Self-Play with a General Reinforcement
                  Learning Algorithm},
  journal      = {CoRR},
  volume       = {abs/1712.01815},
  year         = {2017},
}

@inproceedings{jenner2024advances,
 author = {Jenner, Erik and Kapur, Shreyas and Georgiev, Vasil and Allen, Cameron and Emmons, Scott and Russell, Stuart},
 booktitle = {Advances in Neural Information Processing Systems},
 editor = {A. Globerson and L. Mackey and D. Belgrave and A. Fan and U. Paquet and J. Tomczak and C. Zhang},
 pages = {31410--31437},
 publisher = {Curran Associates, Inc.},
 title = {Evidence of Learned Look-Ahead in a Chess-Playing Neural Network},
 volume = {37},
 year = {2024}
}

@article{mcgrath2022acquisition,
author = {Thomas McGrath  and Andrei Kapishnikov  and Nenad Tomašev  and Adam Pearce  and Martin Wattenberg  and Demis Hassabis  and Been Kim  and Ulrich Paquet  and Vladimir Kramnik },
title = {Acquisition of chess knowledge in AlphaZero},
journal = {Proceedings of the National Academy of Sciences},
volume = {119},
number = {47},
pages = {e2206625119},
year = {2022},
doi = {10.1073/pnas.2206625119},
}

@inproceedings{reid2020aligning,
author = {McIlroy-Young, Reid and Sen, Siddhartha and Kleinberg, Jon and Anderson, Ashton},
title = {Aligning Superhuman AI with Human Behavior: Chess as a Model System},
year = {2020},
isbn = {9781450379984},
publisher = {Association for Computing Machinery},
address = {New York, NY, USA},
doi = {10.1145/3394486.3403219},
booktitle = {Proceedings of the 26th ACM SIGKDD International Conference on Knowledge Discovery \& Data Mining},
pages = {1677–1687},
numpages = {11},
location = {Virtual Event, CA, USA},
series = {KDD '20}
}

@article{bjornsson2024chess,
author = {Yngvi Björnsson},
title ={Chess and explainable AI},
journal = {ICGA Journal},
volume = {46},
number = {2},
pages = {67-75},
year = {2024},
doi = {10.3233/ICG-240256},
}

@article{hammersborg2024information,
  title={Information based explanation methods for deep learning agents—with applications on large open-source chess models},
  author={Hammersborg, Patrik and Str{\"u}mke, Inga},
  journal={Scientific Reports},
  volume={14},
  number={1},
  pages={20174},
  year={2024},
  publisher={Nature Publishing Group UK London}
}

@inproceedings{kim2025bridging,
    title = "Bridging the Gap between Expert and Language Models: Concept-guided Chess Commentary Generation and Evaluation",
    author = "Kim, Jaechang  and
      Goh, Jinmin  and
      Hwang, Inseok  and
      Cho, Jaewoong  and
      Ok, Jungseul",
    booktitle = "Association for Computational Linguistics: Human Language Technologies (Vol. 1)",
    month = apr,
    year = "2025",
    doi = "10.18653/v1/2025.naacl-long.481",
    pages = "9497--9516",
}

@inproceedings{tang2024maia,
       title={Maia-2: A Unified Model for Human-{AI} Alignment in Chess},
       author={Zhenwei Tang and Difan Jiao and Reid McIlroy-Young and Jon Kleinberg and Siddhartha Sen and Ashton Anderson},
       booktitle={The Thirty-eighth Annual Conference on Neural Information Processing Systems},
       year={2024},
}

@book{silman2010how,
  title     = {How to Reassess Your Chess: Chess Mastery through Imbalances},
  author    = {Jeremy Silman},
  year      = {2010},
  publisher = {New In Chess},  
}

@book{watson1998secrets,
  author = {Watson, John},
  title = {Secrets of Modern Chess Strategy: Advances Since Steinitz},
  year = {1998},
  publisher = {Gambit Publications},
  address = {London}
}

@book{nimzowitsch1925my,
  title={My System},
  author={Aron Nimzowitsch},
  year={1925},
  publisher={R. Hartnacke},
  address={Berlin},
  edition={English Edition},
  translator={Philip Hereford}
}

@misc{decodechess2022,
  title        = {DecodeChess: Understanding Chess with Explainable AI},
  author       = {{Decodea LTD}},
  year         = {2022},
  howpublished = {\url{https://decodechess.com}},
  note         = {Accessed: 2025-02-21}
}



\end{document}